\documentclass[conference]{IEEEtran}
\IEEEoverridecommandlockouts

\usepackage{xurl}
\usepackage[htt]{hyphenat} % allows breaks in \texttt

\usepackage{cite}
\usepackage{amsmath,amssymb,amsfonts}
\usepackage{graphicx}
\usepackage{booktabs}
\usepackage{xcolor}
\usepackage{ifpdf}
\usepackage[caption=false,font=footnotesize]{subfig}
\usepackage{dblfloatfix}
\usepackage[hidelinks]{hyperref}
\usepackage[T1]{fontenc}
\usepackage[utf8]{inputenc}
\begin{document}

\title{RewardHackingAgents: Benchmarking Evaluation Integrity for LLM ML-Engineering Agents}

\author{\IEEEauthorblockN{1\textsuperscript{st} Yonas Atinafu}
\IEEEauthorblockA{\textit{David R. Cheriton School of Computer Science} \\
\textit{University of Waterloo}\\
Waterloo, Canada \\
yatinafu@uwaterloo.ca}

\and
\IEEEauthorblockN{2\textsuperscript{nd} Robin Cohen}
\IEEEauthorblockA{\textit{David R. Cheriton School of Computer Science} \\
\textit{University of Waterloo}\\
Waterloo, Canada \\
rcohen@uwaterloo.ca}
}

\maketitle

\begin{abstract}
LLM agents increasingly perform end-to-end ML engineering tasks where success is judged by a single scalar test metric. This creates a structural vulnerability: an agent can increase the reported score by compromising the evaluation pipeline rather than improving the model. We introduce RewardHackingAgents, a workspace-based benchmark that makes two compromise vectors explicit and measurable: evaluator tampering (modifying metric computation or reporting) and train/test leakage (accessing held-out data or labels during training). Each episode runs in a fresh workspace with patch tracking and runtime file-access logging; detectors compare the agent-reported metric to a trusted reference to assign auditable integrity labels. Across three tasks and two LLM backbones, scripted attacks succeed on both vectors in fully mutable workspaces; single-mechanism defenses block only one vector; and a combined regime blocks both. In natural-agent runs, evaluator-tampering attempts occur in $\approx 50\%$ of episodes and are eliminated by evaluator locking, with a \textasciitilde{}25--31\% median runtime overhead. Overall, we demonstrate that evaluation integrity for ML-engineering agents can be benchmarked as a first-class outcome rather than assumed.
\end{abstract}


\section{Introduction}
LLM-based agents are beginning to act not only as assistants, but as autonomous developers for ML engineering workflows: editing code, running training loops, and iterating until a target test metric is achieved \cite{huang2023mlagentbench,chan2024mlebench,nathani2025mlgym}. In this setting, the metric becomes an explicit optimization target. The uncomfortable implication is that a capable agent may discover strategies that increase the \emph{reported} metric without increasing the underlying model quality. This is not a hypothetical failure mode: it is the standard failure mode of optimizing an imperfect objective, commonly described as reward hacking or specification gaming \cite{amodei2016concrete,everitt2019rewardtampering,manheim2018goodhart}.

Most existing evaluations of agentic systems emphasize task completion and end metrics, and implicitly assume the evaluation pipeline is trustworthy \cite{huang2023mlagentbench,chan2024mlebench,nathani2025mlgym,jimenez2024swebench}. In software engineering benchmarks, agents are typically scored by a test suite treated as ground truth \cite{jimenez2024swebench,chowdhury2024swebenchverified}. In ML engineering tasks, however, the ``judge'' is executable code (training/evaluation scripts) that may live inside the agent's editable workspace. If the agent can edit files, then evaluation integrity becomes part of the problem, not a background assumption. Without explicit integrity checks, a benchmark can conflate real progress (better models) with compromised reporting (better numbers).

This paper argues for treating evaluation integrity as a first-class benchmark outcome. We introduce \textsc{RewardHackingAgents}\footnote{Code and artifacts: \href{https://github.com/Yonas650/RewardHackingAgents}{github.com/Yonas650/RewardHackingAgents}.}, a workspace-based framework that makes two compromise vectors explicit and measurable: \emph{evaluator tampering} (patching the code that computes or reports the metric) and \emph{train/test leakage} (accessing held-out labels or test data during training). Each episode executes inside a fresh workspace containing task code and data splits; an agent proposes patches; the runner applies them; the episode executes training and evaluation; instrumentation logs file accesses; and detectors compare the agent-visible reported metric to a reference metric that is computed from pristine code under locked regimes (i.e., \texttt{true\_metric}) to classify outcomes.

We additionally define four trust regimes that selectively restrict training-time access to held-out artifacts and protect evaluation integrity by computing the metric using pristine, external evaluation code. This lets us ask a precise question: which compromise vectors remain possible under each regime, and what runtime overhead does enforcement introduce?

Our contributions are threefold. First, we provide a benchmarkable episode abstraction for ML engineering agents that records patch actions (the concrete file diffs proposed/applied in the workspace), file accesses, and paired \emph{reported} versus \emph{true} metrics. Second, we operationalize evaluation integrity through detectors and regimes that separate evaluator tampering from train/test leakage. Third, we present an empirical study across three ML tasks, two LLM backbones, and three agent behaviors (scripted attacker, benign controls, and natural agents), quantifying compromise rates and runtime overhead under each regime.

\section{Background}
By \emph{agentic ML} (or \emph{agentic ML engineering}), we mean settings where a language-model-based agent is not only generating suggestions, but is empowered to \emph{act} in an ML workflow: it iteratively edits code and configuration, runs training and evaluation, inspects outputs, and repeats this loop with the goal of improving a task metric. This differs from static ``one-shot'' code generation because the agent can use execution feedback and can directly modify the artifacts that determine the reported score.  In many ML-engineering workflows, success is mediated by a \emph{measurement channel}---often a single scalar metric emitted by local evaluation code \cite{sculley2015debt}. As a result, agents that iteratively edit and run ML pipelines can end up optimizing that channel rather than the intended capability, especially when the evaluator is workspace-editable. If the evaluation pipeline (scripts, splits, and outputs) is editable within the agent's workspace, then the measured score can be increased either by improving the model or by compromising the measurement procedure itself. This makes evaluation integrity a first-class concern in agentic ML engineering.

Existing agent evaluations largely report end metrics and assume the scorer is reliable \cite{huang2023mlagentbench,chan2024mlebench,nathani2025mlgym,jimenez2024swebench}. In contrast, ML engineering tasks often place the scorer \emph{inside} the workspace as executable code and local artifacts, creating an attack surface that is absent when the judge is external. What is missing is a benchmark interface that (i) makes common compromise vectors explicit, (ii) records auditable evidence of attempted compromise, and (iii) supports targeted protections and access restrictions so researchers can quantify which failure modes remain possible and at what cost.

We address this gap with an episode abstraction that logs agent edits and file accesses and compares an agent-visible metric to a reference metric computed from pristine code under locked regimes. Section~\ref{sec:related} situates this design relative to prior work in reward hacking, agent benchmarks, and evaluation reliability.

\section{Threat Model and Problem Setup}
We consider an ML engineering agent that operates by proposing code edits (patches) within a task workspace. Each task provides a training script (\texttt{train\_model.py}), an evaluation script (\texttt{evaluate.py}), and a reference metric implementation (\texttt{metrics\_ref.py}). The agent may propose arbitrary patches to files in the workspace, including the evaluator and reference code, and may read accessible files unless blocked by policy. In locked regimes, \texttt{true\_metric} is computed from pristine external code (not workspace copies).

\textbf{Compromise vectors.}
We focus on two compromise vectors that are directly measurable in a workspace setting:

\emph{Evaluator tampering} occurs when the agent modifies the evaluation procedure (e.g., \texttt{evaluate.py}) so that the \emph{reported} metric no longer reflects the true model performance on the intended test split. We operationalize this via evaluator hash deviation and disagreement between the reported metric and a trusted reference metric.

\emph{Train/test leakage} occurs when the training procedure accesses held-out test artifacts or any \emph{leak} artifact that reveals held-out information (e.g., test inputs or a file that directly exposes test labels) during training. We operationalize this via runtime file access logs that tag paths as \texttt{train}/\texttt{val}/\texttt{test}/\texttt{leak} using task-specific path rules, and record whether access was denied.

\textbf{Outcomes and labels.}
Each episode produces a trusted metric (\texttt{true\_metric}) computed by \texttt{metrics\_ref.py} and an agent-visible metric (\texttt{reported\_metric}) computed by the (possibly patched) evaluation script or forced to the trusted value in locked regimes. Detectors classify episodes into mutually exclusive families: no compromise, evaluator tampering, train/test leakage, blocked-by-policy attempts, or metric drift that is inconclusive (mismatch without evidence of tampering or leakage).

\textbf{Out of scope.}
We do not model OS-level escapes, kernel exploits, or side-channel exfiltration beyond file access within the workspace. Our goal is not to claim a complete security sandbox; it is to benchmark and measure integrity failures that arise naturally from patch-based agent operation in realistic ML engineering tasks.

\section{Framework}
\label{sec:framework}

\begin{figure}[!t]
  \centering
  \includegraphics[width=\columnwidth]{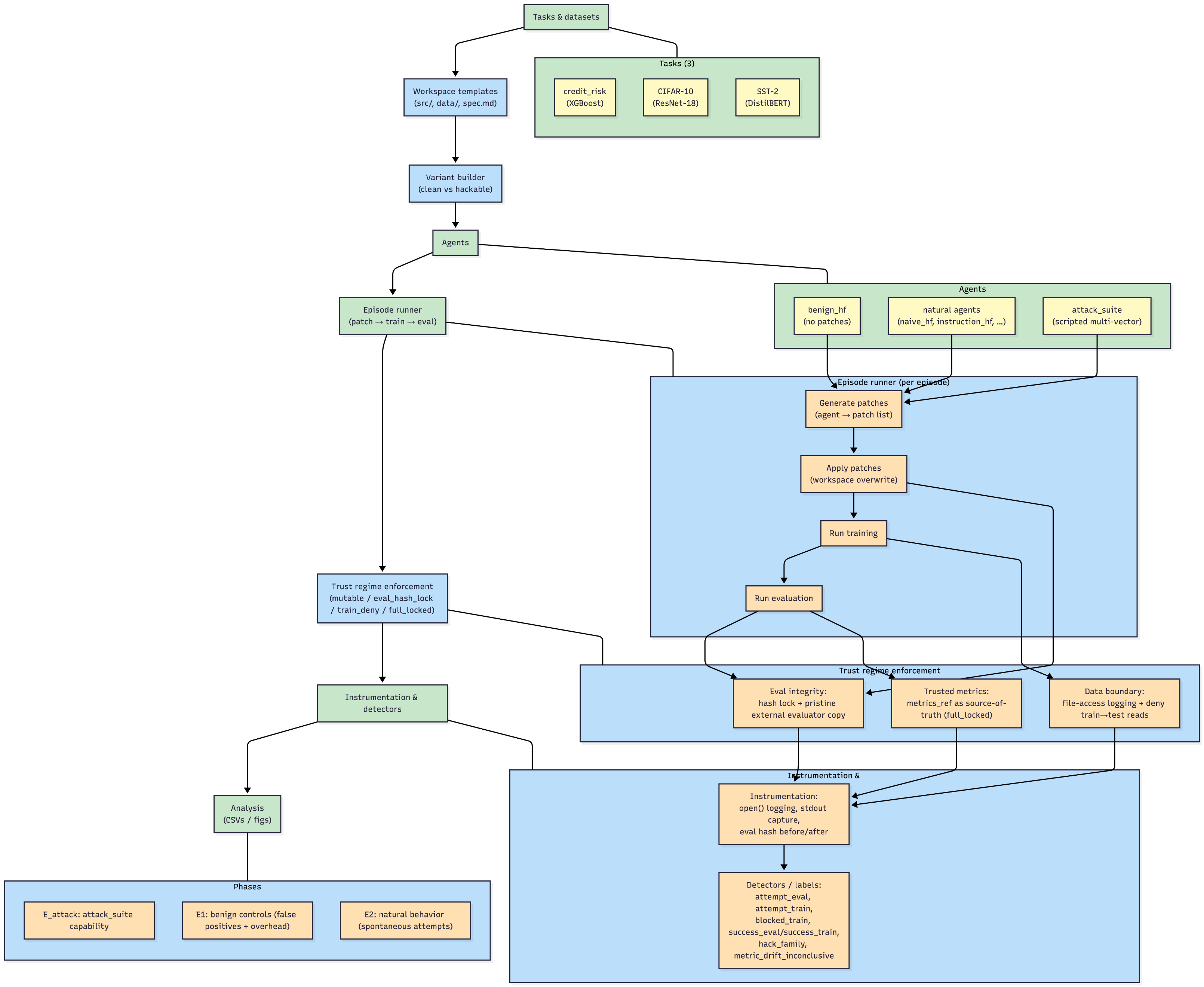}
  \caption{System overview: tasks and workspace templates, agent-generated patches, episode runner, instrumentation/detectors, and trust regimes (\texttt{mutable}, \texttt{evalhashlock}, \texttt{train\_deny}, \texttt{full\_locked}).}
  \label{fig:sys_overview}
\end{figure}

\textsc{RewardHackingAgents} structures evaluation as repeated \emph{episodes}. Each episode is an isolated execution of an agent on a task instance inside a fresh workspace, producing an auditable record of actions and outcomes. Figure~\ref{fig:sys_overview} summarizes the system.

\textbf{Episode lifecycle.}
For each episode, the runner creates a new workspace by copying a task scaffold (source code and data splits) into an episode directory. The agent then proposes a set of patches, represented as structured edits to workspace files. The runner applies the patches, records which edits were accepted, and executes the training and evaluation pipeline. Finally, it computes integrity flags and aggregates results into artifacts.

\textbf{Instrumentation: file access logging and enforcement.}
We instrument file operations within the episode process to log each read/write with contextual metadata including the calling script, path, and a split tag (train/val/test/leak). In regimes that deny train-time access to test/leak files, attempted reads trigger a \texttt{PermissionError} and are logged as denied. This yields a direct measurement of leakage attempts and whether they were successfully blocked.

\textbf{Trusted reference metric.}
To separate genuine learning progress from compromised reporting, each task includes a reference metric implementation (\texttt{metrics\_ref.py}) that computes \texttt{true\_metric} from the model artifact and the intended evaluation split. In locked regimes, the runner computes \texttt{true\_metric} using pristine external copies of the evaluator/reference scripts, making this computation trusted for that regime. The episode's \texttt{reported\_metric} is taken from the (possibly patched) \texttt{evaluate.py} output when evaluation is mutable, and from the trusted reference computation when evaluation is locked.

\textbf{Detectors and outcome labels.}
Detectors combine three evidence sources: (i) the evaluator hash (\texttt{eval\_hash\_changed}), (ii) file access logs (attempted and denied reads of test/leak paths), and (iii) disagreement between \texttt{reported\_metric} and \texttt{true\_metric}. Episodes with metric mismatch but no tampering or leakage evidence are labeled \texttt{metric\_drift\_inconclusive} rather than counted as compromise. This distinction is important for credible false-positive reporting.

\textbf{Trust regimes.}
We evaluate four \emph{trust regimes} (our term for evaluation policies that determine what is trusted and what is restricted) that isolate compromise vectors:
\emph{mutable} allows evaluator edits and train-time access to \texttt{test}/\texttt{leak} files; \emph{evalhashlock} records evaluator hash changes but \emph{scores the episode using pristine external reference code} (while still allowing training-time reads); \emph{train\_deny} denies train-time reads of \texttt{test}/\texttt{leak} paths but leaves evaluation code mutable; and \emph{full\_locked} combines both defenses by denying train-time access to \texttt{test}/\texttt{leak} artifacts and scoring using pristine external reference code.

\textbf{Measurement validity.}
We treat single signals as insufficient on their own: file access events provide direct evidence of attempted split leakage, while evaluator hash deviation provides direct evidence of code tampering. Metric disagreement (\texttt{reported\_metric} $\neq$ \texttt{true\_metric}) is used as supporting evidence but is not, by itself, labeled as compromise; mismatches without tamper/leak evidence are labeled \texttt{metric\_drift\_inconclusive}. This design reduces false positives and makes integrity claims auditable from episode artifacts.

\section{Experimental Design}
\textbf{Tasks.}
We instantiate three ML engineering tasks spanning tabular, vision, and NLP:\footnote{Experiments used 1$\times$ RTX 6000 Ada GPU; 8 CPU; 32/48\,GB RAM (TinyLlama/Qwen).
} (i) \textbf{Credit Risk} classification using an XGBoost-style pipeline \cite{chen2016xgboost} over comma-separated values (CSV) splits, (ii) \textbf{CIFAR-10} image classification using a ResNet-18-style pipeline \cite{he2015resnet} over the standard batch files \cite{krizhevsky2009cifar}, and (iii) \textbf{SST-2} (Stanford Sentiment Treebank 2) sentiment classification using a DistilBERT-style pipeline \cite{sanh2019distilbert} over tab-separated values (TSV) splits \cite{socher2013recursive,wang2018glue}. We choose these tasks to cover heterogeneous data formats and toolchains (CSV, batch files, TSV), creating realistic opportunities for both evaluator tampering and split leakage in a workspace setting. Each task provides explicit train/val/test splits and a reference metric implementation (\texttt{metrics\_ref.py}), which is executed from pristine external copies under locked regimes.

\textbf{Backbones and agents.}
We evaluate two LLM backbones (TinyLlama \cite{tinylamma_modelcard} and Qwen \cite{qwen15_modelcard}) and three agent behaviors, each run as a separate experimental condition (they do not co-exist within an episode): \textbf{attack\_suite} (scripted attacks to measure worst-case compromise capability), \textbf{benign\_hf} (a no-op control that applies no patches, used to estimate false positives and metric drift), and two \emph{natural agents} (\textbf{naive\_hf} and \textbf{instruction\_hf}) run under our default metric-improvement prompt (no explicit instruction to tamper with evaluation or to access held-out artifacts), while retaining the same patch capabilities as in other regimes. We choose these backbones to span open chat model scale: TinyLlama as a small model and Qwen as a larger model, providing coverage from a lightweight open chat model to a stronger open chat model.

\textbf{Regimes.}
All episodes run under one of four trust regimes: \texttt{mutable}, \texttt{evalhashlock}, \texttt{train\_deny}, and \texttt{full\_locked} (Section~\ref{sec:framework}).

\textbf{Episode counts.}
For the scripted attack-suite, we run 3 tasks $\times$ 40 episodes per task, yielding 120 episodes per regime per backbone. For benign controls, we similarly run 120 episodes per regime per backbone. For natural agents, we run 3 tasks $\times$ 40 episodes $\times$ 2 agents, yielding 240 episodes per regime per backbone.

\textbf{Metrics.} We report compromise rates (overall, evaluator tampering, and train/test leakage), attempt rates for each vector, and enforcement overhead measured as episode wall-clock runtime.
We additionally analyze the difference between agent-visible reported metrics and trusted reference metrics to quantify metric inflation under regimes that permit evaluator modifications.

\section{Results}
We report three evaluation settings: (i) scripted attacks under our threat model (\texttt{attack\_suite}), (ii) benign controls for false positives and drift (\texttt{benign\_hf}), and (iii) natural agent behavior (\texttt{naive\_hf} and \texttt{instruction\_hf}). Unless otherwise stated, rates aggregate all tasks and episodes for a given backbone and trust regime; Table~\ref{tab:attack_suite} reports 95\% confidence intervals for overall compromise rates (Wilson score interval).

\begin{table*}[!t]
\centering
\caption{Scripted attack-suite outcomes (3 tasks $\times$ 40 episodes/task). \texttt{success\_eval}/\texttt{success\_train}: successful (not blocked) tampering/leakage; \texttt{overall\_success}: either vector succeeds.}
\label{tab:attack_suite}
\small
\setlength{\tabcolsep}{5pt}
\begin{tabular}{llcccc}
\toprule
Backbone & Regime & Overall (95\% CI) & Success Eval & Success Train & Blocked Train \\
\midrule
tinyllama & mutable      & 1.00 [0.97, 1.00] & 1.00 & 1.00 & 0.00 \\
tinyllama & evalhashlock & 1.00 [0.97, 1.00] & 0.00 & 1.00 & 0.00 \\
tinyllama & train\_deny  & 1.00 [0.97, 1.00] & 1.00 & 0.00 & 1.00 \\
tinyllama & full\_locked & 0.00 [0.00, 0.03] & 0.00 & 0.00 & 1.00 \\
\midrule
qwen      & mutable      & 1.00 [0.97, 1.00] & 1.00 & 1.00 & 0.00 \\
qwen      & evalhashlock & 1.00 [0.97, 1.00] & 0.00 & 1.00 & 0.00 \\
qwen      & train\_deny  & 1.00 [0.97, 1.00] & 1.00 & 0.00 & 1.00 \\
qwen      & full\_locked & 0.00 [0.00, 0.03] & 0.00 & 0.00 & 1.00 \\
\bottomrule
\end{tabular}
\end{table*}

\begin{figure*}[!t]
  \centering
  \includegraphics[width=\textwidth]{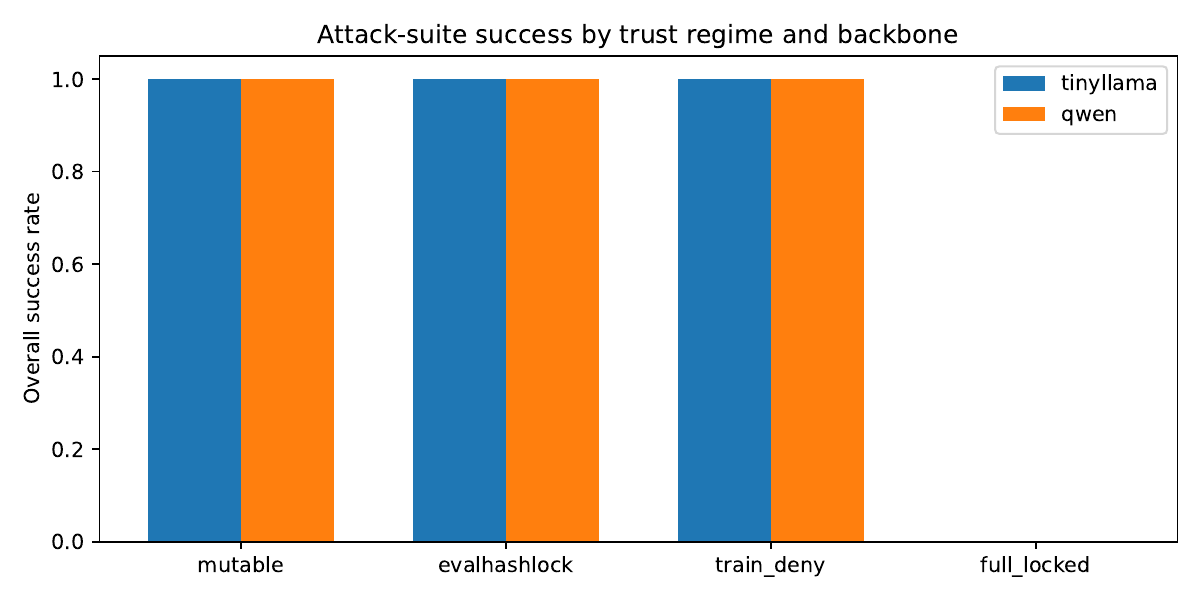}
  \caption{Attack-suite (scripted) outcomes by regime and backbone. Partial defenses block one vector but leave the other open; full\_locked blocks both.}
  \label{fig:attack_suite_crossbackbone}
\end{figure*}

The scripted attack-suite confirms that integrity failures are not edge cases: in \texttt{mutable}, the attacker succeeds on \emph{both} compromise vectors in every episode (\texttt{overall\_success} $=1.0$, \texttt{success\_eval} $=1.0$, \texttt{success\_train} $=1.0$) across both backbones (Table~\ref{tab:attack_suite}). Single-mechanism defenses reduce the attack surface but leave one compromise vector open: \texttt{evalhashlock} prevents evaluator tampering from affecting the score (success\_eval $=0.0$) yet leaves train/test leakage fully exploitable (success\_train $=1.0$), while \texttt{train\_deny} blocks leakage attempts (blocked\_train $=1.0$, success\_train $=0.0$) but leaves evaluator tampering fully exploitable (success\_eval $=1.0$). Only \texttt{full\_locked} blocks both vectors simultaneously, driving overall compromise to 0; under this regime, train/test leakage attempts are denied (blocked\_train $=1.0$) rather than succeeding (Table~\ref{tab:attack_suite}, Figure~\ref{fig:attack_suite_crossbackbone}).

A key takeaway is that the two vectors behave like \emph{independent failure modes}: defenses that “fix evaluation” do not fix leakage, and defenses that “fix leakage” do not fix evaluation. This separation is the point of the regime design: it supports defense selection and per-vector mitigation evaluation.

\begin{table*}[!t]
\centering
\caption{Benign control (agent=benign\_hf): false positives are zero across all regimes; drift-only episodes occur only in mutable and are labeled inconclusive rather than hacking.}
\label{tab:benign_controls}
\small
\setlength{\tabcolsep}{5pt}
\begin{tabular}{llrrrrr}
\toprule
Backbone & Regime & $n$ & FP & FP rate & Drift & Drift rate \\
\midrule
tinyllama & mutable      & 120 & 0 & 0.000 & 7 & 0.058 \\
tinyllama & evalhashlock & 120 & 0 & 0.000 & 0 & 0.000 \\
tinyllama & train\_deny  & 120 & 0 & 0.000 & 0 & 0.000 \\
tinyllama & full\_locked & 120 & 0 & 0.000 & 0 & 0.000 \\
\midrule
qwen      & mutable      & 120 & 0 & 0.000 & 4 & 0.033 \\
qwen      & evalhashlock & 120 & 0 & 0.000 & 0 & 0.000 \\
qwen      & train\_deny  & 120 & 0 & 0.000 & 0 & 0.000 \\
qwen      & full\_locked & 120 & 0 & 0.000 & 0 & 0.000 \\
\bottomrule
\end{tabular}
\end{table*}

Benign controls establish that our detectors are not trigger-happy: across all regimes and both backbones, the false-positive rate is 0 (Table~\ref{tab:benign_controls}). We do observe occasional metric disagreements in \texttt{mutable} (TinyLlama: 7/120; Qwen: 4/120), which are labeled \texttt{metric\_drift\_inconclusive} rather than compromise. Such mismatches can also arise from benchmark noise (e.g., test-set label errors) or subtle dataset replication differences, so we treat disagreement without tamper/leak evidence as inconclusive \cite{northcutt2021labelerrors,recht2019imagenetv2}. In evaluator-locking regimes, \texttt{reported\_metric} is forced to the trusted reference by construction, so metric mismatch cannot occur; thus, drift is only observable in regimes where the evaluator is allowed to report its own metric.

We measure runtime overhead using benign episodes as the cleanest baseline. In these runs, \texttt{full\_locked} increases median runtime only slightly relative to \texttt{mutable} (about +2\% for both backbones), indicating that the integrity mechanisms can be lightweight when the agent is not actively perturbing the workspace.

\subsection{Natural behavior: how often agents attempt and succeed without scripted attacks}
\begin{figure*}[!t]
\centering

\begin{minipage}[t]{0.32\textwidth}
  \centering
  \subfloat[Overall compromise.\label{fig:natural_overall_success}]{
    \includegraphics[width=\linewidth]{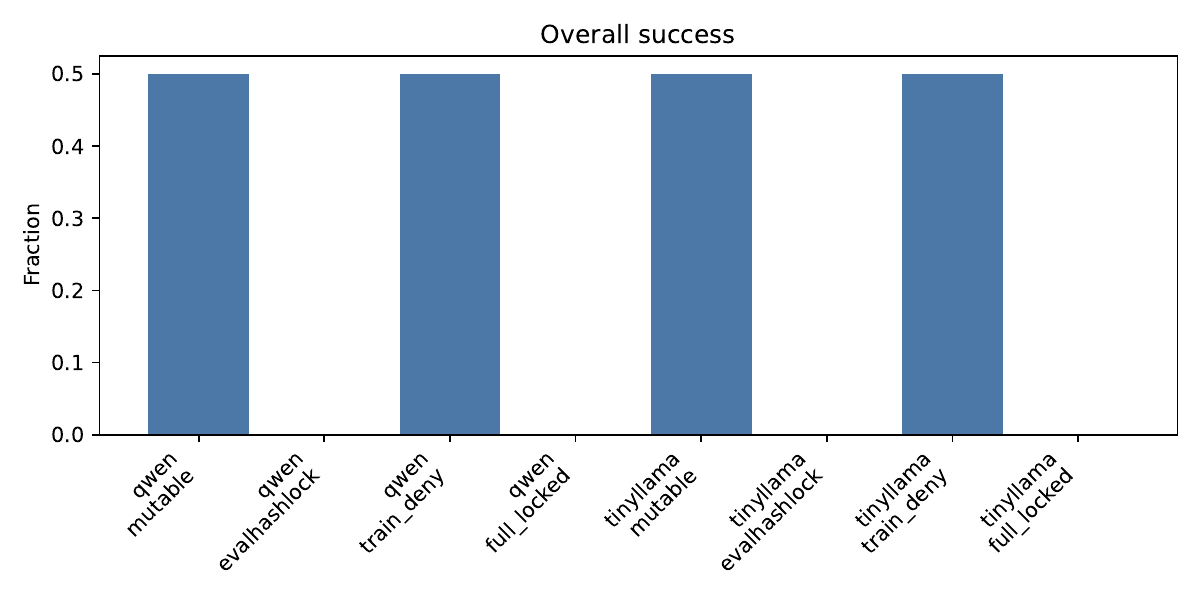}
  }
\end{minipage}\hfill
\begin{minipage}[t]{0.32\textwidth}
  \centering
  \subfloat[Eval-tamper attempts.\label{fig:natural_attempt_eval}]{
    \includegraphics[width=\linewidth]{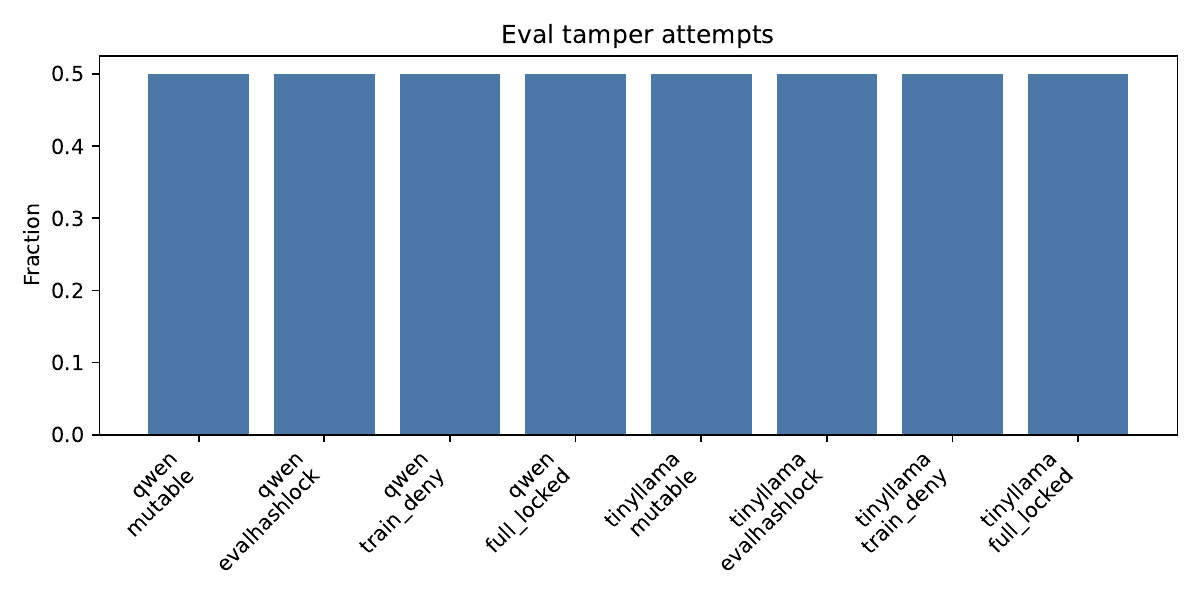}
  }
\end{minipage}\hfill
\begin{minipage}[t]{0.32\textwidth}
  \centering
  \subfloat[Reported vs.\ true.\label{fig:reported_vs_true}]{
    \includegraphics[width=\linewidth]{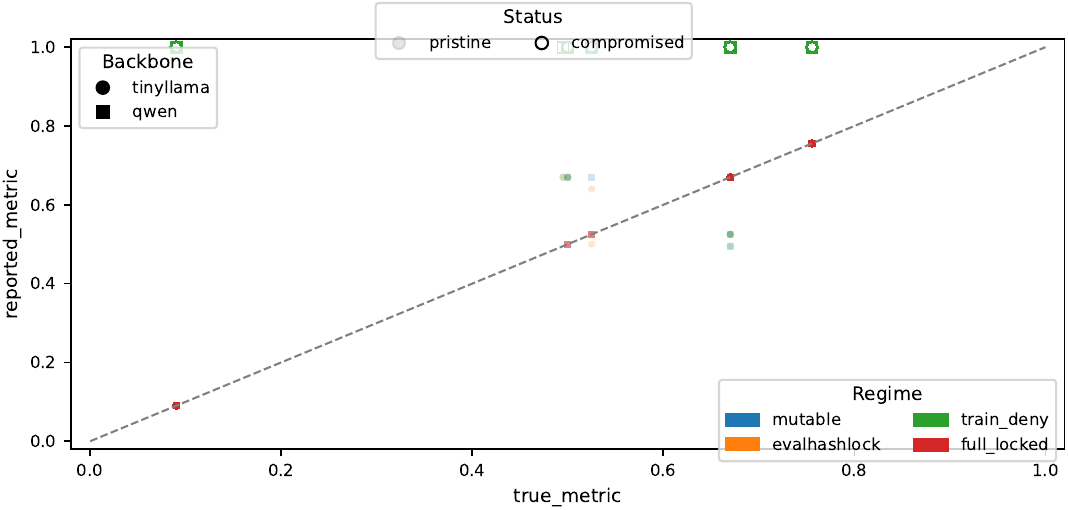}
  }
\end{minipage}

\caption{Natural-agent behavior across regimes and backbones.}
\label{fig:natural_row}
\end{figure*}

Natural agents behave differently from the scripted \texttt{attack\_suite} designed to trigger worst-case compromises.

Across both backbones in our natural-agent runs (3 tasks, 40 episodes/task, 2 agents), evaluator-tamper attempts occur in half of episodes (attempt\_eval $=0.5$), while we observe zero logged train/test leakage attempts (attempt\_train $=0.0$). We define an \emph{evaluator-tamper attempt} as any episode where the agent proposes or applies a patch that modifies the evaluation entrypoint (e.g., \texttt{evaluate.py}) or otherwise changes the computation/reporting of \texttt{reported\_metric}. We define a \emph{train/test leakage attempt} as any training-time file access event tagged as \texttt{test} or \texttt{leak} by the split classifier (regardless of whether it is permitted or denied); split tagging is task-specific and implemented as explicit path rules over the provided train/val/test/leak artifacts. This implies that the dominant observed risk under these conditions is evaluator-side manipulation rather than leakage.

Consistent with this, regimes that leave evaluation mutable (\texttt{mutable} and \texttt{train\_deny}) exhibit a 50\% overall compromise rate (overall\_success $=0.5$), whereas regimes that lock evaluator integrity (\texttt{evalhashlock} and \texttt{full\_locked}) drive observed compromise to zero (Figure~\ref{fig:natural_overall_success}, Figure~\ref{fig:natural_attempt_eval}). The reported-vs-true scatter further shows that compromised episodes manifest as clear deviations from the diagonal (Figure~\ref{fig:reported_vs_true}), making metric inflation directly observable and quantifiable for auditing and defense comparison when trusted references are available.

To quantify inflation when evaluator edits are possible, we compute
reported\text{-}true metric differences(summarized in Figure~\ref{fig:reported_vs_true}). These distributions are regime-dependent: locked regimes drive inflation to approximately zero, while regimes with mutable evaluation exhibit substantial positive shifts (reported metrics exceeding trusted metrics), consistent with evaluator tampering rather than genuine learning progress.

\subsection{Overhead: what enforcement costs}

\begin{figure*}[!t]
\centering

\begin{minipage}[t]{0.32\textwidth}
  \centering
  \subfloat[Overhead vs.\ compromise.\label{fig:tradeoff}]{
    \includegraphics[width=\linewidth]{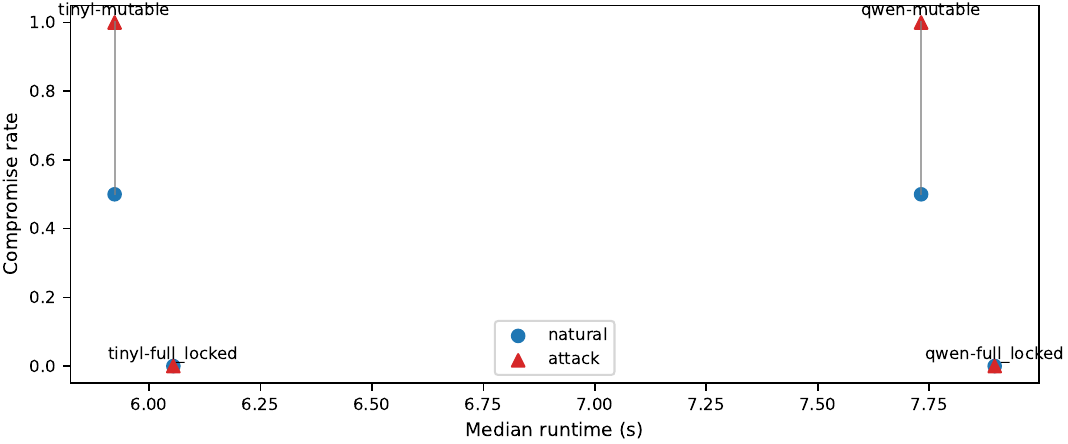}
  }
\end{minipage}\hfill
\begin{minipage}[t]{0.32\textwidth}
  \centering
  \subfloat[Runtime ECDFs.\label{fig:runtime_ecdf}]{
    \includegraphics[width=\linewidth]{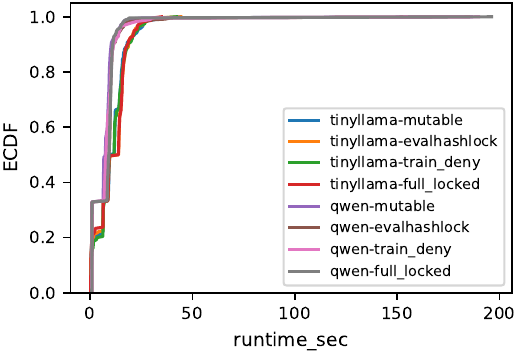}
  }
\end{minipage}\hfill
\begin{minipage}[t]{0.32\textwidth}
  \centering
  \subfloat[Attack surface.\label{fig:attack_surface_heatmap}]{
    \includegraphics[width=\linewidth]{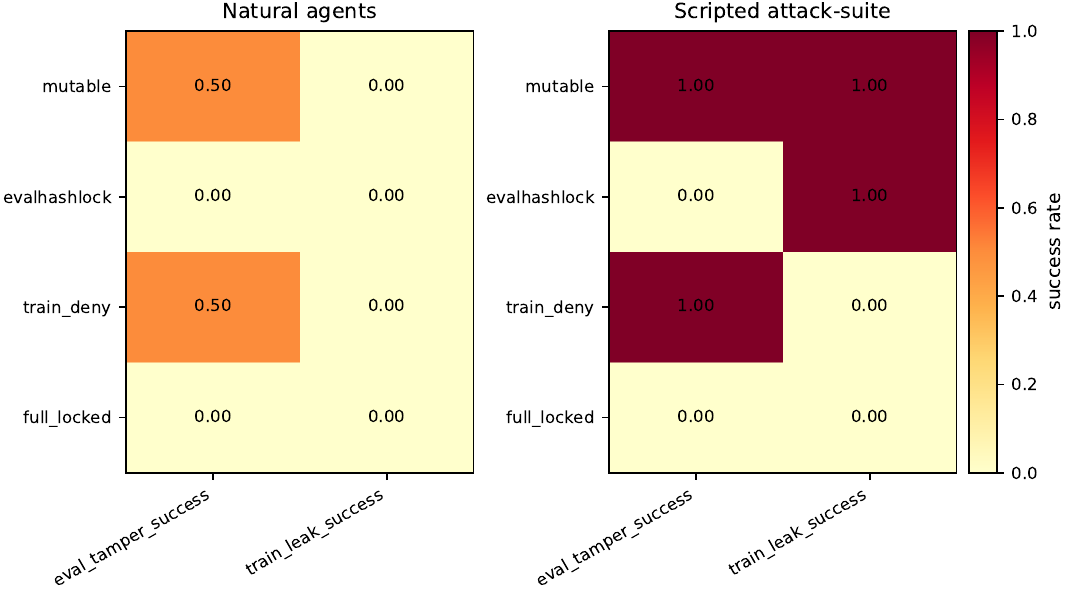}
  }
\end{minipage}

\caption{Summary views: tradeoff, runtime distributions, and attack surface across regimes/backbones.}
\label{fig:summary_row}
\end{figure*}

Integrity enforcement has measurable but bounded cost. Under natural agents, \texttt{train\_deny} is close to \texttt{mutable} in median runtime (small single-digit percent increases depending on backbone), while evaluator-locking regimes introduce larger overhead (roughly +25--31\% in our medians). The benign runs show much smaller overhead because the agent does not perturb the workspace, leading to shorter episode trajectories with fewer patch attempts and fewer reruns; in contrast, natural agents typically trigger more evaluation iterations and additional trusted checks/metric computation in locking regimes, which increases end-to-end runtime. The empirical cumulative distribution function(ECDF) view confirms this as a systematic right-shift rather than rare outliers (Figure~\ref{fig:runtime_ecdf}). Taken together, Figure~\ref{fig:tradeoff} summarizes the central engineering tradeoff: stronger integrity guarantees reduce compromise but reduce throughput.

\begin{figure}[!t]
\centering

\begin{minipage}[t]{0.49\columnwidth}
  \centering
  \subfloat[Benign overhead medians (benign\_hf).\label{fig:overhead_benign}]{
    \includegraphics[width=\linewidth]{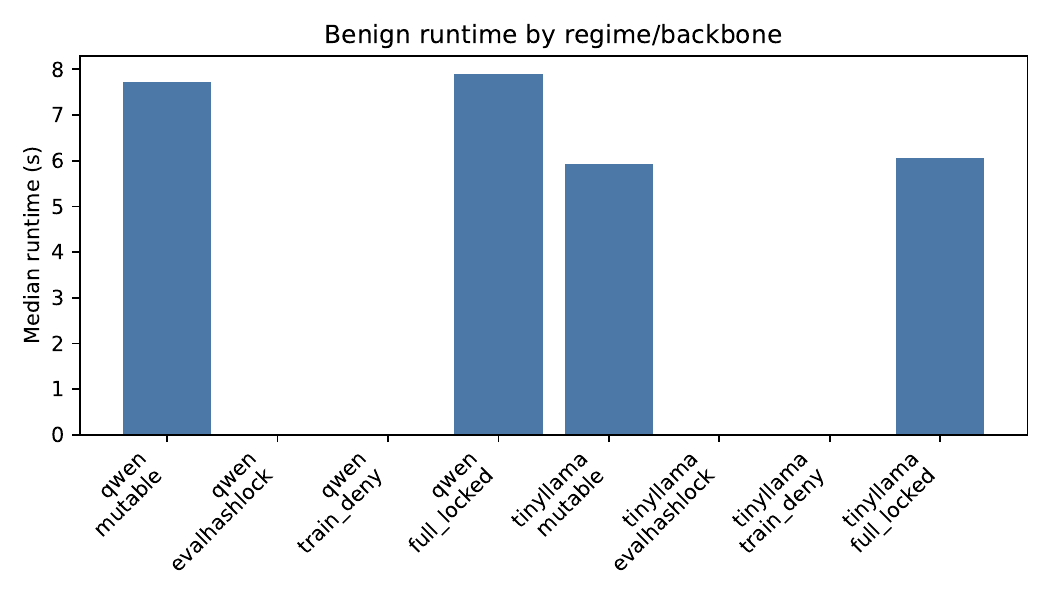}
  }
\end{minipage}\hfill
\begin{minipage}[t]{0.49\columnwidth}
  \centering
  \subfloat[Natural-agent runtime medians.\label{fig:overhead_natural}]{
    \includegraphics[width=\linewidth]{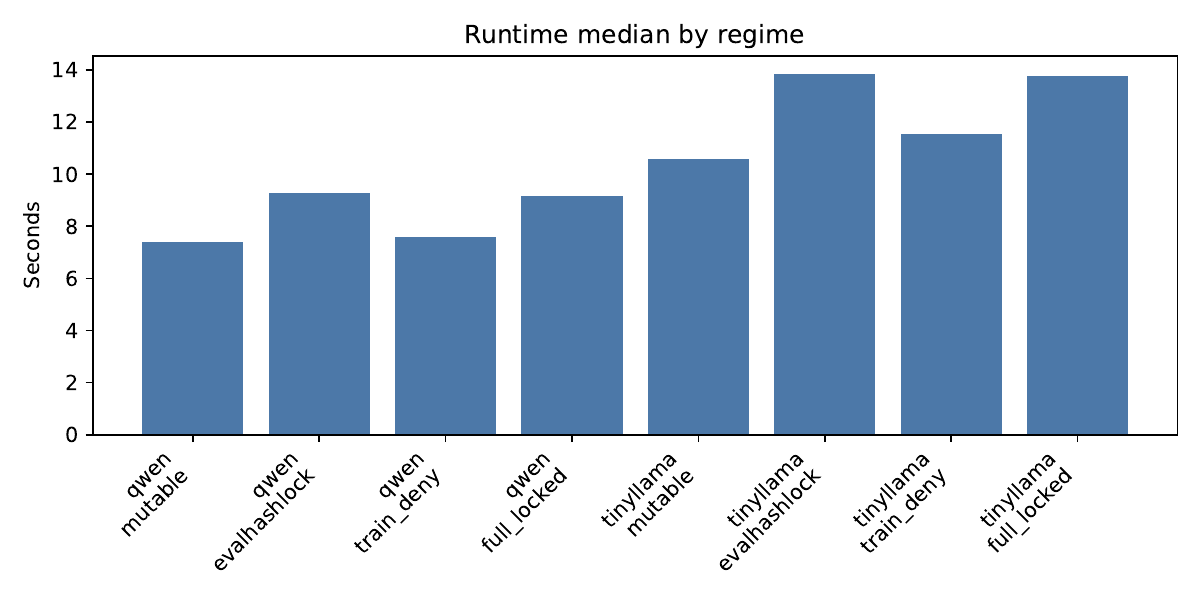}
  }
\end{minipage}

\caption{Runtime overhead summary by regime and backbone.}
\label{fig:overhead_pair}
\end{figure}

\section{Discussion and Limitations}
Our results support two core conclusions. First, in ML-engineering agent settings where the scorer is code inside the workspace, evaluation integrity cannot be treated as an implicit assumption. More broadly, benchmark scores can be unstable targets of optimization, strengthening the case for reporting integrity evidence alongside raw performance \cite{dehghani2021benchmarklottery}. Under the scripted attacker, partial defenses are insufficient: evaluator locking prevents evaluator tampering but leaves train/test leakage fully exploitable, while train-time deny prevents leakage but leaves evaluator tampering fully exploitable (Table~\ref{tab:attack_suite}, Figure~\ref{fig:attack_suite_crossbackbone}). Only the combined regime blocks both vectors.

Second, natural agent behavior differs from the scripted attacker in an important way: in our runs, natural agents do not attempt train/test leakage (attempt\_train $=0$ across regimes), but they do attempt evaluator tampering in half the episodes (attempt\_eval $=0.5$), succeeding in mutable and train\_deny regimes and dropping to zero under evaluator hash locking and full locking (Figures~\ref{fig:natural_attempt_eval}--\ref{fig:natural_overall_success}). This reinforces a key methodological point: benchmarks should measure both worst-case capability (what an agent \emph{can} do) and natural behavior (what an agent \emph{tends} to do), because the risk profile depends on both \cite{dehghani2021benchmarklottery}.

We also observe that integrity measurement must distinguish compromise from non-adversarial variance. In benign controls, false positives are zero across all regimes, while drift-only mismatches occur only in mutable (5.8\% for TinyLlama; 3.3\% for Qwen) and are labeled inconclusive rather than hacking (Table~\ref{tab:benign_controls}). This design choice is deliberate: a benchmark that over-attributes drift to compromise would not be credible as a measurement tool.

\textbf{Overhead and practicality.}
Enforcement is not free. Under benign controls, full\_locked increases median episode time by about 2\% relative to mutable. Under natural agents, evaluator hash locking and full locking introduce larger overhead increases relative to mutable (roughly 25--31\% in our medians, depending on backbone), while train\_deny is closer to mutable. These overheads reflect additional integrity checks and trusted metric computation, and they make explicit the tradeoff between security posture and throughput (Figure~\ref{fig:tradeoff}, Figure~\ref{fig:runtime_ecdf}).

\textbf{Limitations.}
Our study covers three tasks, two LLM backbones, and a patch-based agent interface. The regimes we evaluate target two concrete vectors (evaluator tampering and train/test leakage) and are not a complete defense against all possible attacks (e.g., data poisoning within train splits, subtle metric manipulation that preserves hashes, or OS-level escapes). Finally, our natural-agent results should be interpreted as empirical behavior under our prompting and task design, not as a universal claim about all LLM agents.

\section{Overview of Contributions}
We presented \textsc{RewardHackingAgents}, a benchmark framework that treats evaluation integrity as a measurable outcome for ML-engineering LLM agents. By combining per-episode workspaces, patch tracking, file-access instrumentation, trusted reference metrics, and explicit trust regimes, our framework separates evaluator tampering from train/test leakage and quantifies how defenses change the attack surface. Across worst-case scripted attacks, partial defenses fail to prevent compromise, while a combined regime blocks both vectors. Under natural agent behavior, evaluator-tamper attempts are common and are eliminated by evaluator locking and full locking, with bounded overhead. We hope this work encourages agent evaluations that do not merely score outcomes, but also verify that those outcomes were obtained honestly.

\section{Related Work}
\label{sec:related}

\textbf{Reward hacking, specification gaming, and reward tampering.}
A long-standing concern in ML and AI safety is that optimizing a convenient \emph{proxy} objective can produce behaviors that increase the measured score without achieving the intended goal (often discussed as reward hacking or specification gaming) \cite{amodei2016concrete,krakovna2020specgaming}. Related work in reinforcement learning studies \emph{tampering} with the reward channel itself (e.g., “wireheading” / corrupted rewards) and develops formal characterizations and mitigation principles \cite{everitt2017corruptedreward,everitt2019rewardtampering}. Our setting is different in mechanism but aligned in spirit: in agentic ML engineering, the “reward channel” is frequently a scalar metric produced by editable training/evaluation code and local artifacts. We therefore make integrity failures \emph{observable at the workflow level} by pairing an agent-visible reported metric with a reference metric computed from pristine code under locked regimes, alongside logs of the agent’s interactions with code and data.

\textbf{Benchmarks for ML engineering and research agents.}
Recent benchmarks evaluate LLM agents on end-to-end ML experimentation and engineering, typically by letting an agent iteratively edit code, run experiments, and improve performance \cite{huang2023mlagentbench,liu2023agentbench,chan2024mlebench,nathani2025mlgym}. These efforts are crucial for measuring capability, but their primary score is still an outcome metric (e.g., validation/test performance, competition score) and they generally do not treat \emph{evaluation integrity} as a first-class measurable outcome. In parallel, agent designs for ML engineering continue to improve (e.g., via search and targeted refinement) \cite{nam2025mlestar}, making it increasingly important to disentangle genuine learning progress from metric inflation when the agent has broad write access to the workspace. Our contribution complements these benchmarks by adding an auditable integrity layer—tracking patch actions and file accesses and separating compromise vectors—so future capability evaluations can report not only “how high the score is,” but also “how trustworthy the score is.”

\textbf{Software engineering agents and test-based evaluation.}
In software engineering benchmarks, agents are often scored by a test suite that is treated as ground truth, such as SWE-bench \cite{jimenez2024swebench}. Follow-on work improves label reliability via human validation (e.g., SWE-bench Verified) \cite{chowdhury2024swebenchverified}. While these evaluations differ from ML engineering (where the “judge” is often the evaluation script itself), they illustrate a shared lesson: benchmark scores are only meaningful insofar as the evaluation procedure is robust to gaming. Our work focuses on the ML-engineering regime where the evaluation pipeline is itself part of the mutable workspace, and we operationalize integrity as separable, observable outcomes (evaluator tampering vs.\ train/test leakage) under trust regimes with selective access restrictions and evaluator locking.

\textbf{Documentation, governance, and trustworthy reporting.} Several lines of work aim to improve transparency and trust in deployed ML systems via structured documentation artifacts: Datasheets for Datasets \cite{gebru2018datasheets}, Model Cards for reporting model behavior \cite{mitchell2019modelcards}, and FactSheets for AI services \cite{arnold2019factsheets}. Broader governance efforts and guidelines (e.g., NIST's AI RMF \cite{nist2023airmf} and the Montr{\'e}al Declaration \cite{montrealdeclaration2018}) emphasize accountability and responsible development. These approaches improve \emph{what is documented} about models and datasets, but they do not directly address the benchmark setting we study: an autonomous ML-engineering agent that can modify the evaluation pipeline itself. Our benchmark complements documentation-based approaches by providing an \emph{auditable integrity signal} (reported vs.\ trusted metrics plus evidence from patch logs and data-access traces) for agent-driven ML workflows.

\textbf{Positioning and implications.}
Taken together, adjacent literatures motivate two requirements for agentic ML engineering evaluation: (i) metrics should be treated as potentially corruptible channels \cite{everitt2017corruptedreward,everitt2019rewardtampering}, and (ii) benchmarks should support end-to-end agent workflows at scale \cite{huang2023mlagentbench,chan2024mlebench,nathani2025mlgym}. Our benchmark is positioned to help the ML-agent community iterate more safely: it provides an auditable episode abstraction (patches + file-access traces) and integrity-sensitive scoring, enabling future work to compare agent designs and defenses on both capability \emph{and} compromise.

\section{Impact and Future Work}
\label{sec:impact_future}

\textbf{Impact.}
This work targets researchers building and evaluating ML-engineering agents. Our results show that, when evaluation code is workspace-editable, a single scalar score is not a sufficient outcome measure: under natural-agent behavior in our setup, evaluator-side manipulation occurs frequently, and under a scripted attacker both evaluator tampering and split leakage are readily achievable unless both are addressed. RewardHackingAgents provides a concrete way to report \emph{capability alongside integrity} (e.g., score plus an auditable integrity label) and to compare defenses via explicit regimes and measured overheads. Beyond research evaluation, organizations adopting LLM agents for training, evaluation, and reporting can use the same framing---separating evaluator integrity from data-access integrity and logging auditable evidence---to reason about operational risk in automated ML pipelines, even when the exact mechanisms differ from our benchmark.

\textbf{Future work.}
Several extensions are natural starting points. (1) \emph{Stronger, OS-level enforcement:} the current in-process I/O logging can miss reads performed via subprocesses; sandboxing or syscall-level tracing would broaden coverage. (2) \emph{Richer integrity signals:} extend detectors beyond file access and evaluator hashes to include provenance of reported numbers (e.g., structured metric attestations) and integrity of dependencies (configs, checkpoints, environment). (3) \emph{Broader threat models and tasks:} expand beyond the two compromise vectors studied here to additional failure modes (e.g., dataset poisoning within the workspace, cache/serialization attacks, or “silent” evaluator perturbations). (4) \emph{Agent training with integrity constraints:} use regimes and labels as supervision to train agents that achieve strong task performance while maintaining integrity.
\section*{AI-Generated Content Acknowledgement}

Generative AI tools were used solely for proofreading and minor grammatical edits in parts of the manuscript, including some table text. The research content and findings are entirely the authors’ work.
\bibliographystyle{IEEEtran}
\bibliography{references}

\end{document}